\documentclass[10pt,twocolumn,letterpaper]{article}

\usepackage[pagenumbers]{cvpr} 

\definecolor{cvprblue}{rgb}{0.21,0.49,0.74}
\usepackage[pagebackref,breaklinks,colorlinks,allcolors=cvprblue]{hyperref}
\usepackage{dsfont}
\usepackage{multirow}

\def\confName{CVPR}
\def\confYear{2026}

\title{{\ourmethod}: Portrait Video Camera Control via Scale-Aware Conditioning}

\author{Weijie Lyu$^{1,2*}$ \quad Ming-Hsuan Yang$^1$ \quad Zhixin Shu$^{2 \dag}$\vspace{0.2cm}\\
$^1$University of California, Merced \quad $^2$Adobe Research\vspace{0.2cm}\\
}
\newcommand\blfootnote[1]{%
  \begingroup
  \renewcommand\thefootnote{}\footnote{#1}%
  \addtocounter{footnote}{-1}%
  \endgroup
}

\newcommand{\ourmethod}{\textit{FaceCam}}

\usepackage{algorithm}
\usepackage{algorithmicx}
\usepackage{algpseudocode}  

\begin{document}
\captionsetup{hypcap=false}
\twocolumn[{%
\renewcommand\twocolumn[1][]{#1}%
\maketitle
\begin{center}
\vspace{-8mm}
\centering
\includegraphics[width=0.98\linewidth]{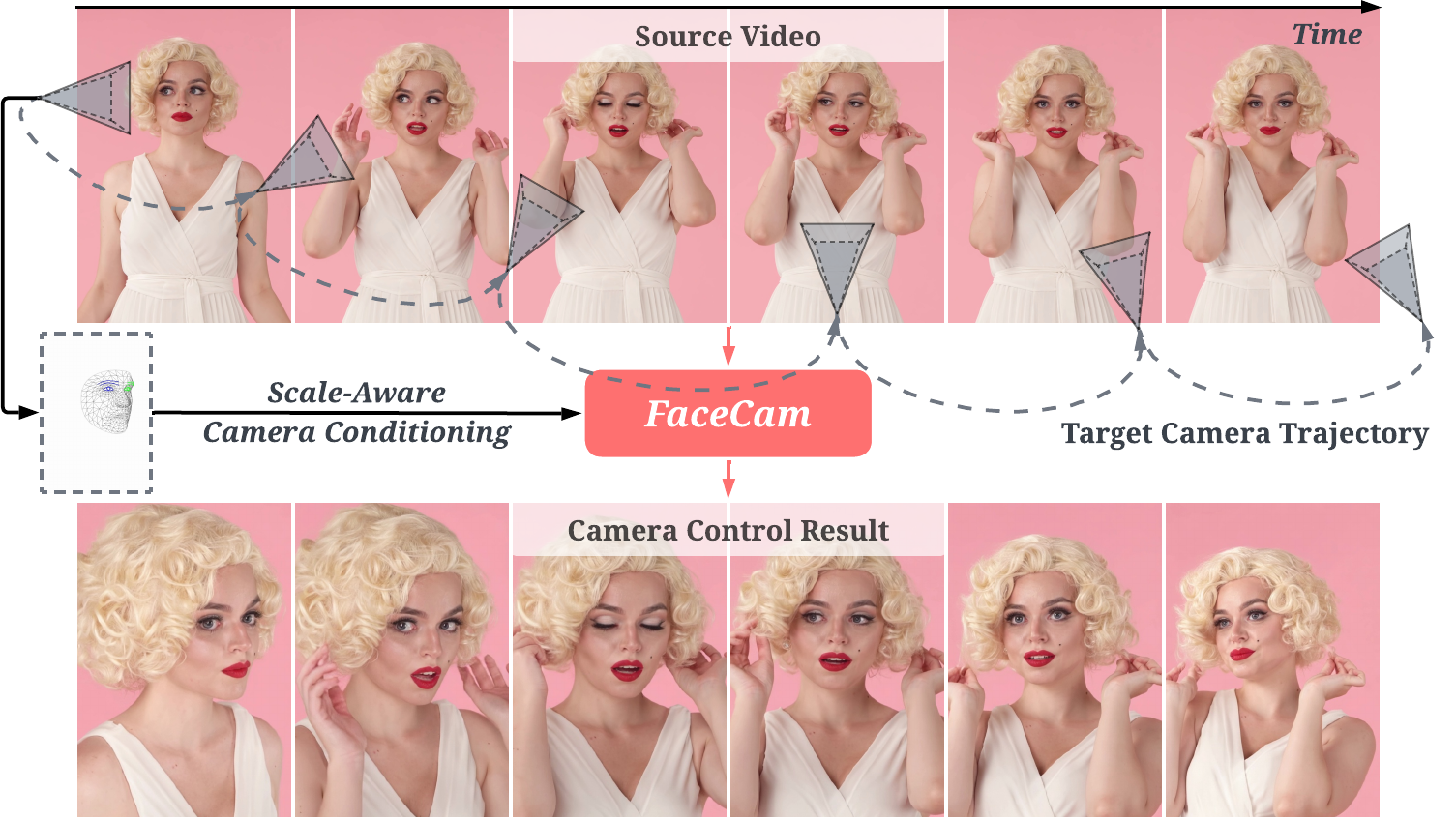}
\captionof{figure}{{\ourmethod} generates portrait videos with precise camera control from a single input video and a target camera trajectory. We introduce \textit{scale-aware camera conditioning} that represents the target camera via rendered facial landmarks, enabling accurate camera pose control. Our approach preserves subject identity and motion while maintaining high visual quality. Project page: \url{https://weijielyu.github.io/FaceCam}.}
\label{fig:teaser}
\end{center}
}]

\blfootnote{*Work was done when Weijie Lyu was an intern at Adobe Research.}
\blfootnote{\dag Corresponding author.}

\vspace{-2mm}
\begin{abstract}
We introduce {\ourmethod}, a system that generates video under customizable camera trajectories for monocular human portrait video input.
Recent camera control approaches based on large video-generation models have shown promising progress but often exhibit geometric distortions and visual artifacts on portrait videos due to scale-ambiguous camera representations or 3D reconstruction errors.
To overcome these limitations, we propose a face-tailored scale-aware representation for camera transformations that provides deterministic conditioning without relying on 3D priors.
We train a video generation model on both multi-view studio captures and in-the-wild monocular videos, and introduce two camera-control data generation strategies: synthetic camera motion and multi-shot stitching, to exploit stationary training cameras while generalizing to dynamic, continuous camera trajectories at inference time.
Experiments on Ava-256 dataset and diverse in-the-wild videos demonstrate that {\ourmethod} achieves superior performance in camera controllability, visual quality, identity and motion preservation.
\end{abstract}    
\vspace{-4mm}
\section{Introduction}
\vspace{-1mm}
\label{sec:intro}
Controllable video generation~\cite{controllable, wan_animate, wan_move, motion_prompting, motionctrl, trajectory_attention} has emerged as a central topic in recent research, with camera motion~\cite{cameractrl, generative_camera_dolly, recapture, ac3d, vd3d} being one of its most critical control dimensions. Meanwhile, human portrait videos are among the most prevalent video formats, making camera control for portraits a key problem in computer vision and graphics, with applications in social media, post-production, telepresence, and AR/VR.
Given a source video, the goal of a camera-control system is to allow users to specify discrete camera positions or continuous camera trajectories and then generate a video of the same scene from those configurations. This defines a dynamic view-synthesis problem, where the model must infer time-varying scene geometry and synthesize unseen pixels from a learned prior. In the context of portrait videos, maintaining accurate facial expressions, verbal articulation, identity consistency, and subtle motions, such as head movement and hair dynamics, is critical for perceptual quality.

Contemporary approaches typically build on large foundation video generation models as strong visual priors. Two main strategies have emerged for specifying camera control. The first~\cite{generative_camera_dolly, recammaster, cameractrl, ac3d, vd3d} employs scene-agnostic camera representations, such as intrinsic and extrinsic parameters or image-like encodings of rays (\eg, Plücker rays~\cite{plucker}). The second~\cite{trajectorycrafter, recapture, geometry_forcing, trajectory_attention} infers camera motion from scene reconstruction, \eg using depth estimation, thereby tying control directly to the underlying 3D structure.

For human portrait videos, existing approaches face notable challenges. Scene-agnostic camera representations, being unaware of video content, make it difficult to specify the desired camera changes for a portrait and suffer from scale ambiguity: the same parameter change can induce dramatically different visual transformations depending on the object or scene scale, shown in Fig.~\ref{fig:scale_amb}. Reconstruction-based methods rely on 3D understanding~\cite{depthcrafter, vggt} to derive camera motion; small geometric errors in these estimates can amplify into large perceptual artifacts, such as shape distortions or identity drift. These artifacts are especially noticeable due to human sensitivity to facial appearance and facial expressions.
The second challenge in training camera control for portrait video generation is data: acquiring paired videos with ground-truth camera annotations that capture the full complexity of human dynamics. Real portrait videos must preserve dynamic facial expressions, natural head movements, and fine-grained details such as realistic hair motion, all of which are notoriously difficult to simulate synthetically at scale. The core difficulty lies in obtaining paired training data where the same dynamic scene is recorded under different camera trajectories.

To address these challenges, we propose {\ourmethod}, a portrait video generation system with precise camera control. We overcome the limitations of existing camera-conditioning schemes by introducing a scale-aware camera representation that encodes the relative transformation between source and target poses using image-space pixel correspondences. By explicitly modeling how camera motion acts on a 3D human head, this representation resolves monocular scale ambiguity and allows users to specify camera trajectories in a more direct and interpretable way.
To enhance the model’s ability to preserve dynamic facial expressions, natural head movements, and fine-grained details, we train our network on NeRSemble~\cite{nersemble}, a studio-captured multi-view human video dataset. However, this dataset only provides static cameras. To enable continuously moving camera trajectories at inference, we introduce two data generation strategies: synthetic camera motion and multi-shot stitching. We find that the discontinuous camera pose changes produced by multi-shot stitching during training generalize well to continuous camera trajectories at inference. We further incorporate in-the-wild videos augmented with synthetic camera motion to mitigate overfitting to the studio lighting conditions.

By leveraging a large video generation model as backbone and initialization for fine-tuning, we achieve state-of-the-art performance with high fidelity across two key dimensions: precise camera control adherence and faithful preservation of subject dynamics, including facial expressions, identity, head motion, and realistic hair movement. We validate our method on both the studio-captured Ava-256~\cite{ava256} dataset, which provides ground-truth multi-view static cameras, and challenging in-the-wild portrait videos, demonstrating superior performance in camera control accuracy and video quality compared to existing methods. Our contributions are summarized as follows:

\begin{itemize}
    \item We propose {\ourmethod}, a portrait video camera-control system with a face-tailored, scale-aware camera representation that resolves the scene-scale ambiguity of traditional camera parameterizations and enables intuitive authoring of camera trajectories.
    \item We develop a data generation and training pipeline that, despite using only static-camera multi-view captures and unlabeled in-the-wild videos for training, supports continuous target camera motion in inference, without relying on any 4D synthetic data.
    \item Extensive experiments on in-the-wild data validate the effectiveness of our approach, demonstrating precise camera-control adherence and faithful preservation of subject dynamics, highlighting its promise for real-world applications.
\end{itemize}
\vspace{-1mm}
\section{Related Work}
\vspace{-1mm}
\subsection{Human Face View Synthesis}
\vspace{-1mm}
View synthesis for human portraits has progressed from 3D Morphable Model (3DMM)-based~\cite{3dmm} mesh reconstruction to NeRF-/Gaussian-based heads and, more recently, diffusion-based generation. Classical 3DMM pipelines estimate per-frame pose and expression on a textured mesh and refine appearance across frames to obtain a drivable avatar~\cite{face2face, deepvideoportraits, deferred, neural_head_avatars}, but they struggle to capture fine-scale appearance, complex hair, and full-head coverage. Dynamic NeRF- and Gaussian-based methods further condition on expression codes or FLAME~\cite{flame} parameters~\cite{nerface, imavatar, headnerf, insta, gaussianavatars, fate}, and subsequent variants improve robustness, rendering quality, and articulation for upper-body and full-head avatars. However, monocular pipelines still report inefficiency, difficulty handling large pose changes and rear-head views, and reliance on per-instance optimization over hundreds or thousands of frames, which limits scalability.
Recent diffusion-based approaches~\cite{wang2024vexpress, chen2025echomimic, cui2024hallo3, jiang2024loopy} and foundation-style avatar models~\cite{chen2025hunyuanavatar, kong2025multitalk, omniavatar, omnihuman} instead condition powerful portrait or video diffusion models on audio, text, or sparse motion cues, and scale to multi-identity, multi-character settings with strong lip-sync and expression control. Nevertheless, the primary focus of these works is audio-driven portrait synthesis under limited camera motion; explicitly controlled novel view synthesis and recapturing for portrait videos, especially from a single monocular recording, remains comparatively underexplored.

\vspace{-1mm}
\subsection{Camera-Control Video Generation}
\vspace{-1mm}
Camera control for text/image-conditioned video generation~\cite{motionctrl, cameractrl, vd3d, ac3d} extends large video diffusion models with explicit 3D camera pose or ray-based embeddings to synthesize videos that follow user-specified trajectories from prompts or single images.
For dynamic novel view synthesis, GCD~\cite{generative_camera_dolly} introduces a camera-controlled video-to-video translation pipeline trained on synthetic videos from Kubric~\cite{kubric}, but it suffers from poor generalization to in-the-wild data due to domain gaps. ReCapture~\cite{recapture} generates an anchor video using multi-view diffusion or point-cloud rendering and then applies masked per-video LoRA~\cite{lora} fine-tuning to re-angle user-provided videos. Methods such as NVS-Solver~\cite{nvssolver} and CAT4D~\cite{cat4d} repurpose pre-trained video or multi-view video diffusion models as zero-shot or multi-view backbones for static and dynamic novel view synthesis under target camera poses. More recently, ReCamMaster~\cite{recammaster} trains a camera-controlled generative re-rendering model on a large synthetic multi-view video dataset rendered with Unreal Engine, and TrajectoryCrafter~\cite{trajectorycrafter} uses a dual-stream diffusion model that fuses point-cloud renders with the source video to achieve precise trajectory control and generative inpainting of occluded regions. However, these methods still struggle on portrait videos camera control due to ambiguous camera representations and geometric estimation errors.
\vspace{-1mm}
\section{Method}
\vspace{-1mm}
\subsection{Problem Setup}
\vspace{-1mm}
Consider a dynamic head as a 4D scene $A$, a video $V$ of $f$ number of frames $\{I_i\}_{i=1}^f\in\mathbb{R}^{f\times h\times w\times c}$ is produced by capturing this scene along a per-frame camera trajectory $C=\{\mathbf{P}_i\}_{i=1}^f$, with each camera pose $\mathbf{P}_i=[\mathbf{R}_i\!\mid\!\mathbf{t}_i]\in\mathbb{R}^{3\times4}$. 
Let $\mathbf{K}\in\mathbb{R}^{3\times 3}$ denote camera intrinsics, we can represent the capture process as rendering the 4D scene $A$:
\begin{equation}
    V \;=\; \mathsf{Render}\big(A;\ C,\ \mathbf{K}\big).
\end{equation}
Given a source video $V^s$ captured under a camera trajectory $C^s$, our goal is to generate a target video $V^t$ under a target camera trajectory $C^t$ which captures the same dynamic scene $A$.
In practice, the source camera trajectory $C^s$ is unobtainable, and our system should be able to estimate that. We represent our task as:
\begin{equation}
    V^t = {\ourmethod}(V^s, C^t).
\end{equation}
\renewcommand\thesubfigure{\Alph{subfigure}}
\begin{figure}[t]
  \centering
  \begin{subfigure}{0.98\linewidth}
    \includegraphics[width=\linewidth]{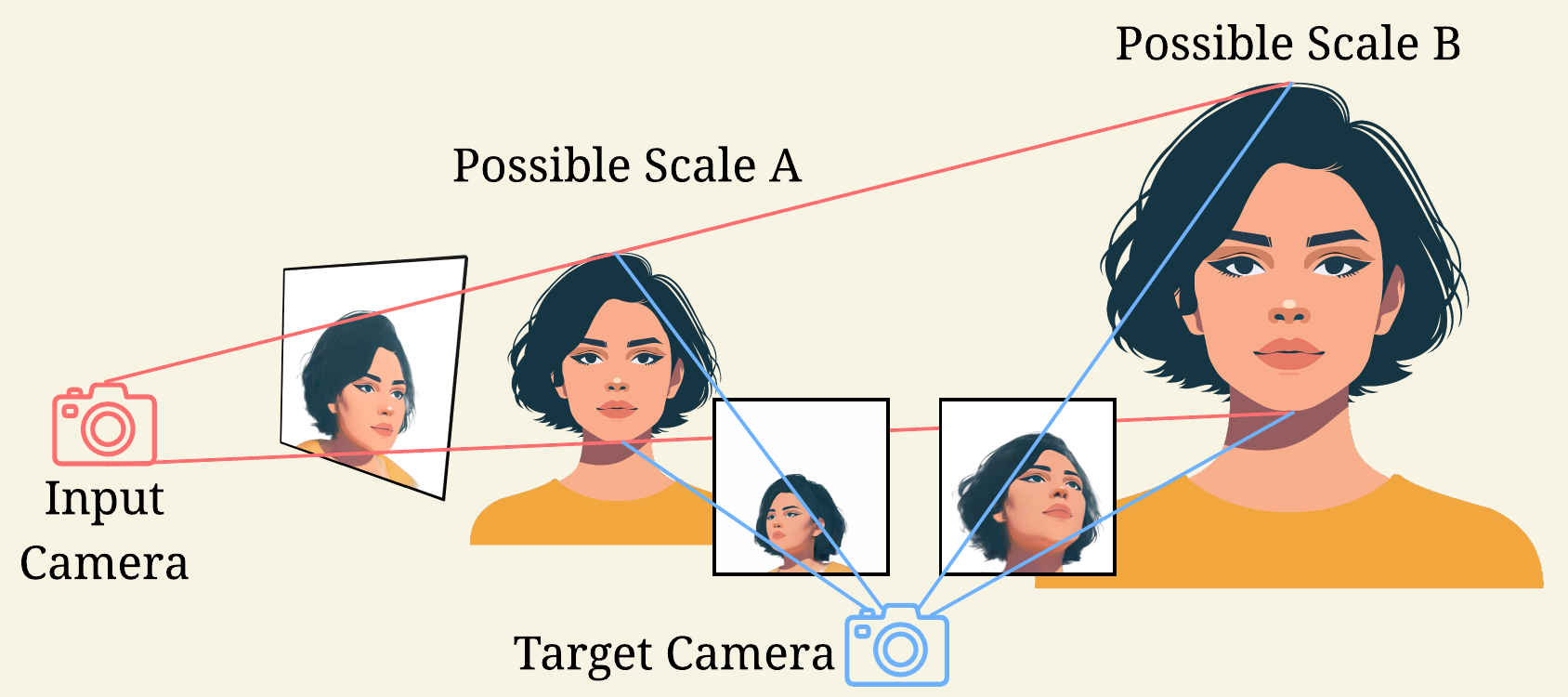}
    \caption{\textbf{Scale-ambiguous camera representation.} Existing camera control methods~\cite{recammaster, cameractrl, ac3d, generative_camera_dolly} encode camera using extrinsic parameters. In monocular capture, metric depth is unobservable, the scene is determined only up to a global similarity with unknown scale and translation. Hence, the same image admits infinitely many 3D configurations, making re-rendering from a target pose underdetermined and leading to drift and poor controllability.}
    \label{fig:scale_amb}
  \end{subfigure}
  \hfill
  \vspace{1mm}
  \begin{subfigure}{0.98\linewidth}
    \includegraphics[width=\linewidth]{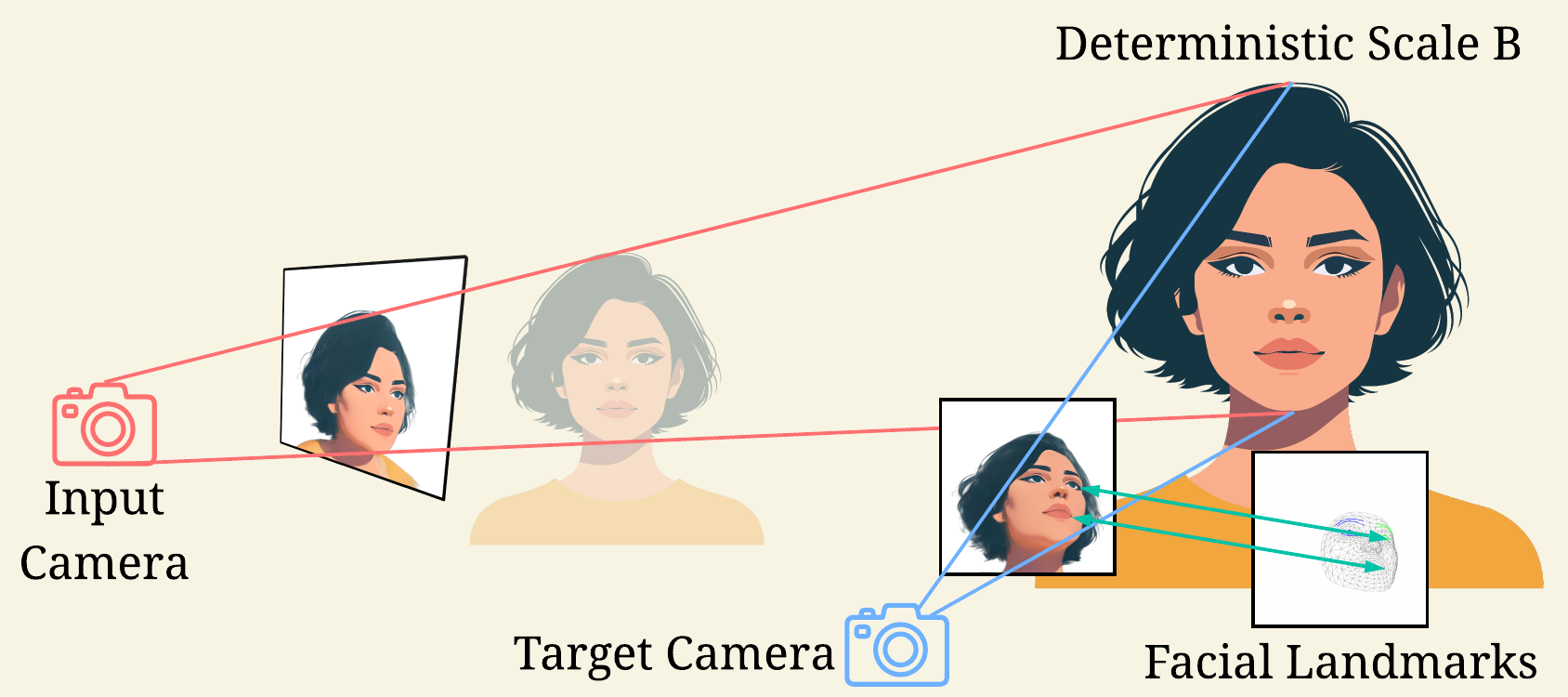}
    \caption{\textbf{Scale-aware camera representation.} Instead of extrinsics, we encode the camera via image-space point correspondences. With at least seven 2D correspondences, the fundamental matrix between two uncalibrated views can be estimated, and with known intrinsics the relative pose is recovered up to a global scale. Portrait videos naturally provide such correspondences through facial landmarks, so we use rasterized 2D landmark maps—renderings of 3D facial landmarks from the anchor frame—as the camera representation. This face-tailored, scale-aware encoding is easy to visualize and enables deterministic, high-precision control of the apparent camera pose.}
    \label{fig:scale_awa}
  \end{subfigure}
  \vspace{-2mm}
  \caption{\textbf{Camera representation comparison.} We contrast (A) parameter-based representations, which are standard in camera control methods, with (B) image-space point correspondences, which we adopt in {\ourmethod} to obtain a scale-aware conditioning that enables precise camera control.}
\label{fig:scale}
\vspace{-2mm}
\end{figure}

\begin{figure*}[t]
  \centering
  \begin{subfigure}[t]{0.40\linewidth}
    \centering
    \includegraphics[width=\linewidth]{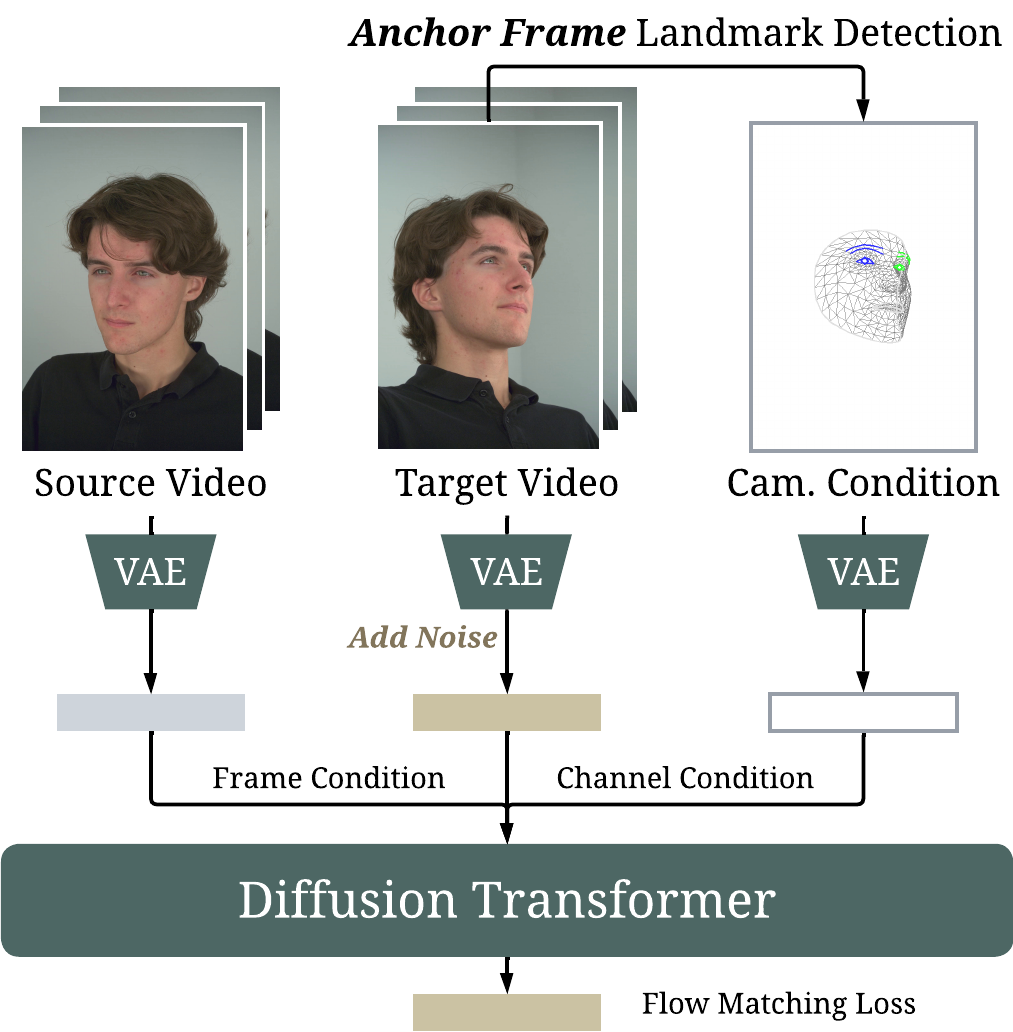}
    \subcaption{\textbf{Training.} We extract facial landmarks from the anchor frame of the target video as camera condition. Source video, target video, and camera condition are encoded by a VAE into latents, which are passed into the diffusion transformer to predict the target latent, optimized with a flow-matching loss.}
    \label{fig:method-train}
  \end{subfigure}\hfill
  \begin{subfigure}[t]{0.57\linewidth}
    \centering
    \includegraphics[width=\linewidth]{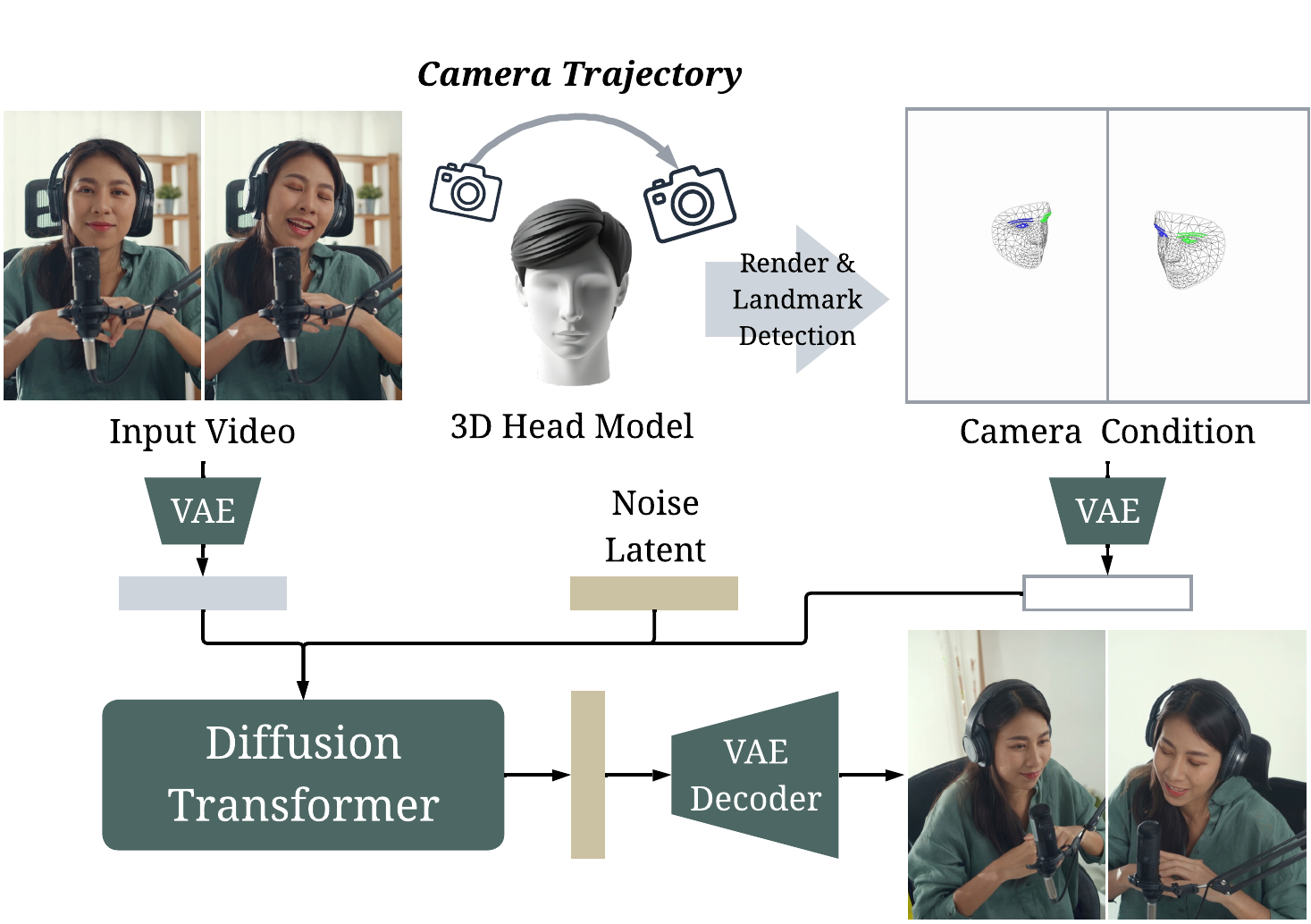}
    \subcaption{\textbf{Inference.} We use a 3D head model generated as a generic head, render it along the target camera trajectory, and detect facial landmarks as the camera condition. The output latent from the diffusion transformer is decoded by a VAE decoder to obtain the camera-controlled video. We observe that, although the model is trained only with discontinuous camera pose changes, it generalizes to continuous camera trajectories during inference.}
    \label{fig:method-infer}
  \end{subfigure}
  \vspace{-2mm}
  \caption{Training and inference pipeline of {\ourmethod}.}
  \label{fig:method}
  \vspace{-4mm}
\end{figure*}
\vspace{-2mm}
\subsection{Camera Representation via Correspondences}
\vspace{-1mm}
\label{subsec:camera}
\noindent\textbf{Image-space correspondences as sufficient camera representation.}
Classical multi-view geometry shows that \emph{image-space point correspondences} are sufficient to characterize relative camera motion. Given two views and a set of corresponding pixels, one can estimate a fundamental matrix $F$ that satisfies the epipolar constraint for each correspondence~\cite{HartleyZisserman2004,Hartley1997}. With known intrinsics $\mathbf{K}$, $F$ is upgraded to the essential matrix $E=\mathbf{K}^\top F \,\mathbf{K}$, from which the relative pose $[\mathbf{R}\!\mid\!\mathbf{t}]$ is recovered up to an unknown global scale by decomposing $E$~\cite{HartleyZisserman2004}. Thus, point correspondences encode exactly the observable camera-induced image formation transform (up to this global scale) and are a sufficient representation of camera motion for control. This representation underpins modern SfM pipelines~\cite{Schonberger2016sfm}: detect repeatable keypoints (\eg, SIFT~\cite{Lowe2004}), match them, robustly estimate $F$ or $E$ with RANSAC~\cite{Fischler1981}, triangulate, and refine with bundle adjustment~\cite{Triggs2000}. Systems like COLMAP~\cite{Schonberger2016sfm} implement this workflow end-to-end and demonstrate its effectiveness at scale.

\vspace{1mm}
\noindent\textbf{Camera parameters and monocular scale ambiguity.}
Many camera control methods~\cite{recammaster, cameractrl, ac3d} directly use camera extrinsics as the conditioning signal. Representing a camera by its extrinsics $\mathbf{P}=[\mathbf{R}\mid\mathbf{t}]$ and intrinsics $\mathbf{K}$ exposes an unobservable degree of freedom. Consider the $i$-th frame of the source video $V^s$, captured under pose $\mathbf{P}^s_i=[\mathbf{R}^s_i\mid \mathbf{t}^s_i]$ towards a dynamic scene $A$ at timestamp $i$:
\begin{equation}
    V^s_i = \mathsf{Render}\big(A_i;\ [\mathbf{R}^s_i\mid \mathbf{t}^s_i],\ \mathbf{K}\big).
\end{equation}
A 3D point $\mathbf{x} \in \mathbb{R}^3$ is expressed in camera coordinates as
\begin{equation}
    \mathbf{x}_c=\mathbf{R}\mathbf{x}+\mathbf{t}=(x_c,y_c,z_c)^\top,
\end{equation}
and projected to pixel coordinates
\begin{equation}
    u=\frac{f_xx_c}{z_c}+c_x,\qquad v=\frac{f_yy_c}{z_c}+c_y,
\end{equation}
where $f_x$, $f_y$, $c_x$, and $c_y$ are from the intrinsic matrix $\mathbf{K}$. Monocular image formation is invariant to a global similarity transform: for any $\alpha>0$, letting $\mathbf{x}'=\alpha\mathbf{x}$ and $\mathbf{t}'=\alpha\mathbf{t}$ yields
\begin{equation}
    \mathbf{x}_c' = \mathbf{R}(\alpha\mathbf{x}) + \alpha\mathbf{t}
    = \alpha(\mathbf{R}\mathbf{x} + \mathbf{t})
    = \alpha\mathbf{x}_c,
\end{equation}
so the perspective ratios $x_c'/z_c'=x_c/z_c$ and $y_c'/z_c'=y_c/z_c$, and hence $(u,v)$, remain unchanged. Given only a single monocular video $V^s$, absolute metric depth and translation magnitude are therefore not observable. A model conditioned directly on $[\mathbf{R}\!\mid\!\mathbf{t}]$ must implicitly choose a metric scale not fixed by the pixels; for a fixed target trajectory $C^t$, this can lead to variation in the 2D placement of the portrait (Fig.~\ref{fig:scale_amb}). In contrast, point correspondences reside in pixel space and never expose this unobservable global scale, as they encode exactly what can be observed.

\subsection{Scale-Aware Camera Conditioning}
\noindent\textbf{Camera conditioning using facial landmarks.}
Facial landmarks provide reliable correspondences for portrait videos. We use these landmarks to implement our correspondence-based camera representation from Sec.~\ref{subsec:camera} (see Fig.~\ref{fig:scale_awa}). We detect $m$ landmarks in the first frame (anchor frame) of the target video and use them to define a head-centric coordinate system. Let $\mathbf{X}=\{\mathbf{x}_k\}_{k=1}^m$ be the 3D positions of these landmarks (from monocular 3D face reconstruction), and let $\mathbf{U}\!=\!\{\mathbf{u}_k\}_{k=1}^m$ be their 2D projections under a desired camera pose $[\mathbf{R}\!\mid\!\mathbf{t}]$ with intrinsics $\mathbf{K}$:
\begin{equation}
    \mathbf{u}_k = (u_k, v_k) = \mathcal{N}\bigl(\mathbf{K}(\mathbf{R}\mathbf{x}_k+\mathbf{t})\bigr),
\end{equation}
where $\mathcal{N}$ performs perspective division: $\mathcal{N}(\mathbf{x}_c) = (x_c/z_c, y_c/z_c)$. We use $\mathbf{U}$ as the camera representation.

\vspace{1mm}
\noindent\textbf{Scale invariance.}
As shown in Sec.~\ref{subsec:camera}, monocular cameras cannot determine absolute scale. Our landmark representation has this same property. If we scale both the 3D landmarks and translation by any factor $s>0$ (\ie, $\mathbf{x}_k' = s\mathbf{x}_k$ and $\mathbf{t}' = s\mathbf{t}$), the 2D projections stay the same:
\begin{equation}
    \mathbf{u}'_k
    = \mathcal{N}\bigl(\mathbf{K}(\mathbf{R}\mathbf{x}'_k+\mathbf{t}')\bigr)
    = \mathcal{N}\bigl(\mathbf{K}s(\mathbf{R}\mathbf{x}_k+\mathbf{t})\bigr)
    = \mathbf{u}_k.
\end{equation}
This shows that $\mathbf{U}$ does not depend on the absolute scale of the scene. Unlike the translation vector $\mathbf{t}$ which is defined in real-world units, $\mathbf{U}$ works directly with what we can observe in pixel space.

\vspace{1mm}
\noindent\textbf{Sufficiency for pose control.}
Our landmark representation contains sufficient information to determine the camera pose. Given 3D landmarks $\mathbf{X}$ and their 2D projections $\mathbf{U}$, a PnP solver can recover the camera rotation $\mathbf{R}$ and translation $\mathbf{t}$ up to a single global scale, since monocular video does not fix absolute distances. This residual scale ambiguity matches our design: both $\mathbf{U}$ and $\mathbf{X}$ are normalized to be scale-invariant. Rather than explicitly solving for pose, we condition the generator directly on rasterized landmark maps.
In practice, rather than feeding 2D landmark coordinates directly as numeric inputs, we rasterize the target landmarks into pixel-space channels and use the resulting images as the conditioning signal. This offers a key practical advantage: users can preview and author the desired camera viewpoint simply by inspecting the rendered facial shape, making camera control intuitive to specify.

\begin{figure}[t]
    \centering
    \includegraphics[width=.98\linewidth]{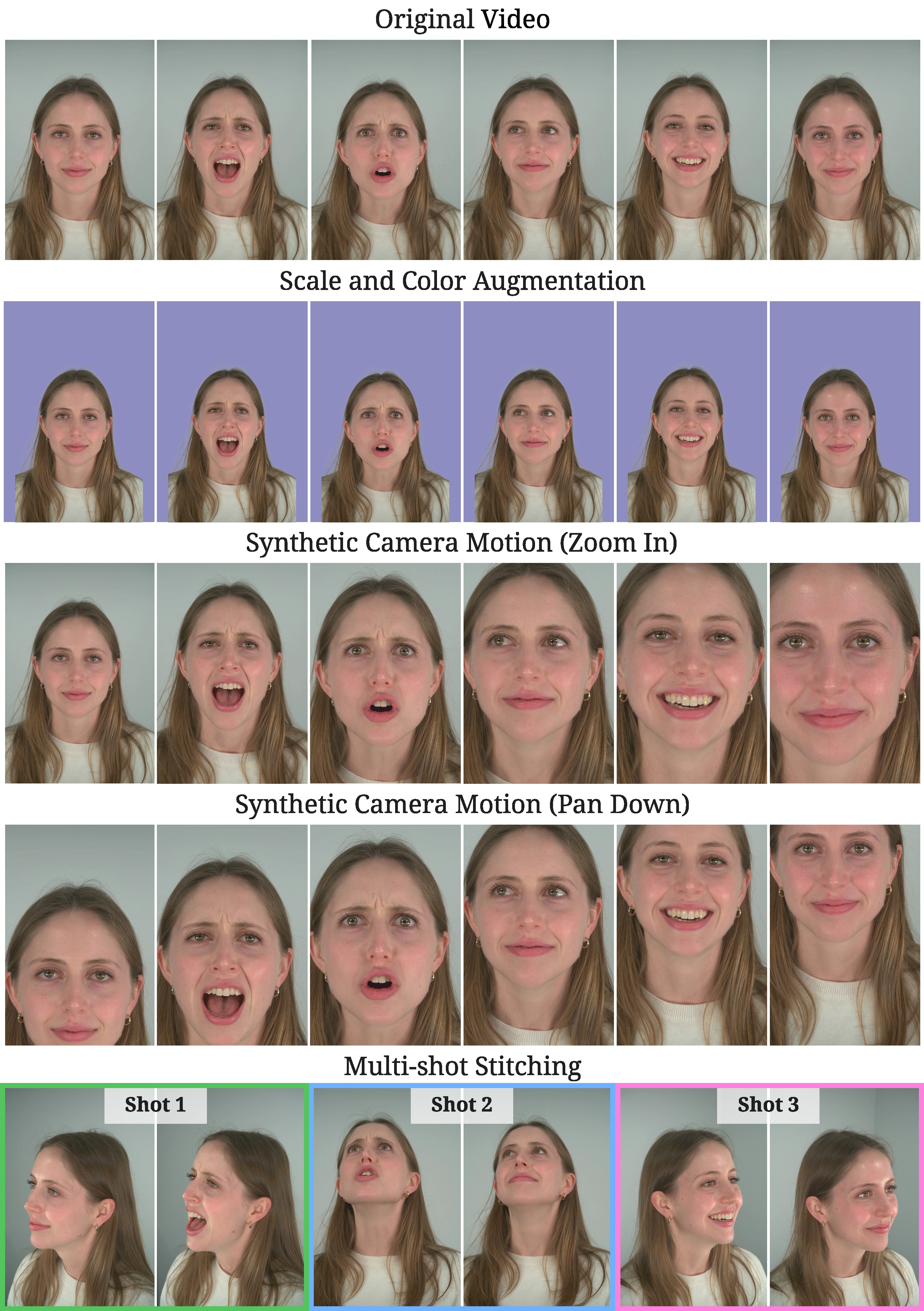}
    \caption{\textbf{Training data generation examples.} The source video is applied with scale and color augmentation to increase data diversity, while the target video is augmented with all three types to train the model’s camera control capability.}
    \label{fig:data_gen}
    \vspace{-4mm}
\end{figure}

\subsection{Training Data Generation}
\label{subsec:data_gen}
A major challenge in implementing dynamic portrait video camera control is acquiring suitable training data.
Scalable synthetic data acquisition for high-quality, realistic 3D dynamic portraits remains challenging; therefore, we rely exclusively on real captured data. We begin with a multi-view video dataset~\cite{nersemble} containing facial performances from 425 subjects captured in a studio environment. Each subject is recorded from 16 synchronized viewpoints with different facial expressions and head movements, yielding approximately 9.4K video sequences. While this dataset provides known camera parameters, it has two critical limitations: first, camera trajectories remain static throughout each sequence, restricting the model to view synthesis between fixed positions. Hence, we develop several data augmentation strategies to synthesize training pairs and enhance model capabilities. Second, all captures share identical studio lighting conditions, limiting the model's ability to generalize to diverse in-the-wild scenarios. To address this limitation, we supplement our training with in-the-wild videos with synthetic camera movement.

\vspace{1mm}
\noindent\textbf{Scale and color augmentation.}
Since NeRSemble~\cite{nersemble} captures faces at uniform scales against fixed studio backgrounds, we introduce variation by (1) randomly scaling each clip with factor $s \in [0.75, 1.25]$, and (2) segmenting the foreground face and replacing the background with random colors (consistent between source and target videos).

\vspace{1mm}
\noindent\textbf{Synthetic camera motion.}
We simulate camera motion on both studio and in-the-wild videos to enable training with dynamic cameras. Specifically, we synthesize two motion types: zoom and pan. For zoom, we sample start and end scale ratios from $[1.0, 1.25]$ and linearly interpolate per frame, producing smooth zoom-in (start $<$ end) or zoom-out (start $>$ end) effects, then restore the original resolution via cropping or padding. For pan, we linearly interpolate cropping or padding offsets across frames, assigning each frame a distinct offset to simulate lateral camera motion parallel to the image plane.

\vspace{1mm}
\noindent\textbf{Multi-shot stitching.}
We can simulate the effect of a moving camera with synthetic camera movement. However, the motion is parallel to the initial image plane and does not include rotation. To introduce camera rotation in model training, we propose a multi-shot stitching technique: for each target video, we randomly select 1–4 clips captured from different camera poses, trim them to different temporal segments, and stitch them together so that a single sequence exhibits changing viewpoints. Although the generated target video only contains discrete camera pose changes, we find through experiments that the model can still perform inference with smooth, continuous camera pose changes.

\vspace{1mm}
\noindent\textbf{Adding in-the-wild data.}
Training our video model exclusively on NeRSemble~\cite{nersemble} suffices to support inference with smoothly varying camera trajectories. However, the generated lighting often deviates from the input video, and hands or accessories can appear malformed due to the dataset’s limited domain. To improve generalization ability, we collect roughly 800 monocular in-the-wild portrait videos. Because these clips lack a second viewpoint, we apply synthetic camera motion to create target videos with virtual camera movement and pair them with the original clips as source videos, thereby diversifying our training set.

\vspace{-1mm}
\subsection{Inference Pipeline}
\label{subsec:inference}
For inference, we offer a user-friendly procedure to generate the camera condition given a target trajectory, as shown in Fig.~\ref{fig:method-infer}. We first take a 3D Gaussian head model produced by FaceLift~\cite{facelift} and render a proxy video along the desired camera trajectory around this head. We then run MediaPipe~\cite{mediapipe} on each rendered frame to obtain a sequence of facial landmarks, which we use as the camera conditioning signal for video generation. Note that the proxy 3D head can be of any identity and is unrelated to the input video, and we use the same 3D Gaussian head model for all experiments. Also, the sequence of facial landmarks is a representation of camera trajectory, not the actual position of the face in generated video, as further illustrated in Fig.~\ref{fig:random}.
\vspace{-1mm}
\section{Experiments}
\vspace{-1mm}
\subsection{Experimental Setup}
\noindent\textbf{Implementation details.}
We build our system on the open-source video foundation model Wan~\cite{wan2025} for conditional video generation.
Following~\cite{recammaster}, we concatenate the source video latent with noise latent through frame condition, and the camera conditioning latent is applied through channel condition following~\cite{wan2025}.
Detailed architecture and training settings are provided in the supplementary material.
We train {\ourmethod} on the dataset introduced in Sec.~\ref{subsec:data_gen}. This dataset contains 8.9K videos generated from NeRSemble~\cite{nersemble}, and about 200 in-the-wild videos, in total about 9.1K videos. We follow~\cite{recammaster} and fine-tune the 3D attention layers and projection layers of the diffusion model. The entire training takes 3K steps on 24 NVIDIA A100 GPUs, with a constant learning rate of 5$e$-5 and a batch size of 24.

\vspace{1mm}
\noindent\textbf{Evaluation datasets and metrics.}
To quantitatively evaluate {\ourmethod}, we construct two benchmarks for camera control video generation.
(1) \textit{Static camera setting}. We select 10 identities from the studio-captured Ava-256 dataset~\cite{ava256}. For each identity, we construct 10 input–output camera pairs, yielding 100 videos. For the baselines, since Ava-256 provides camera extrinsics for each video, we convert these parameters into each method's camera coordinate system to form the camera-control signal. For {\ourmethod}, we detect facial landmarks in the first frame of the target video and use them as the camera conditioning. To ensure a fair comparison when the target video is unavailable, we render a generic 3D Gaussian head under the target camera pose, detect its landmarks, and use them as the conditioning signal; we denote this variant as {\ourmethod}*.
We assess novel view synthesis performance using 
PSNR, SSIM, and LPIPS~\cite{lpips}, and measure identity preservation with ArcFace~\cite{arcface}.
(2) \textit{Dynamic camera setting.} We collect 100 in-the-wild portrait videos to evaluate {\ourmethod} under dynamic camera trajectories. We apply 10 canonical camera motions (Pan Left / Right / Up / Down, Zoom In / Out, Arc Left / Right / Up / Down), following the basic trajectories in~\cite{recammaster}. Each motion is assigned to 10 videos. We evaluate visual quality with VBench~\cite{vbench} and identity preservation with ArcFace~\cite{arcface}.

\vspace{1mm}
\noindent\textbf{Baselines.}
We compare {\ourmethod} with two baselines, ReCamMaster~\cite{recammaster} and TrajectoryCrafter~\cite{trajectorycrafter}, which represent two prevailing strategies for camera control in video generation: a scene-agnostic camera parameters conditioning approach and a reconstruction-based approach. ReCamMaster injects camera extrinsics as conditioning signals into the self-attention layers of DiT~\cite{dit} blocks. TrajectoryCrafter first estimates a dynamic point cloud, renders it under the target camera trajectory, and then achieves camera control by inpainting the rendered dynamic point cloud.

\begin{table}[tb]
  \centering
  \scriptsize
  \caption{\textbf{Quantitative results on Ava-256.} {\ourmethod} outperforms the baselines on both reconstruction metrics and facial identity metric, indicating stronger stationary camera control ability and better preservation of identity and motion.}
  \vspace{-2mm}
  \begin{tabular}{l@{\hskip 5.6mm}c@{\hskip 5.6mm}c@{\hskip 5.6mm}c@{\hskip 5.6mm}c}
    \toprule
    Method & PSNR $\uparrow$ & SSIM $\uparrow$ & LPIPS $\downarrow$ & ArcFace $\uparrow$ \\
    \midrule
    ReCamMaster~\cite{recammaster} & 9.73 & 0.5570 & 0.5809 & 0.7014 \\
    TrajectoryCrafter~\cite{trajectorycrafter} & 10.32 & 0.5462 & 0.5673 & 0.5220\\
    \ourmethod* & 9.83 & 0.5816 & 0.5494 & 0.8073\\
    \ourmethod & \textbf{15.85} & \textbf{0.7208} & \textbf{0.2521} & \textbf{0.8574} \\
  \bottomrule
  \end{tabular}
  \label{tab:ava_256}
  \vspace{-4mm}
\end{table}

\begin{figure}[t]
    \centering
    \includegraphics[width=.98\linewidth]{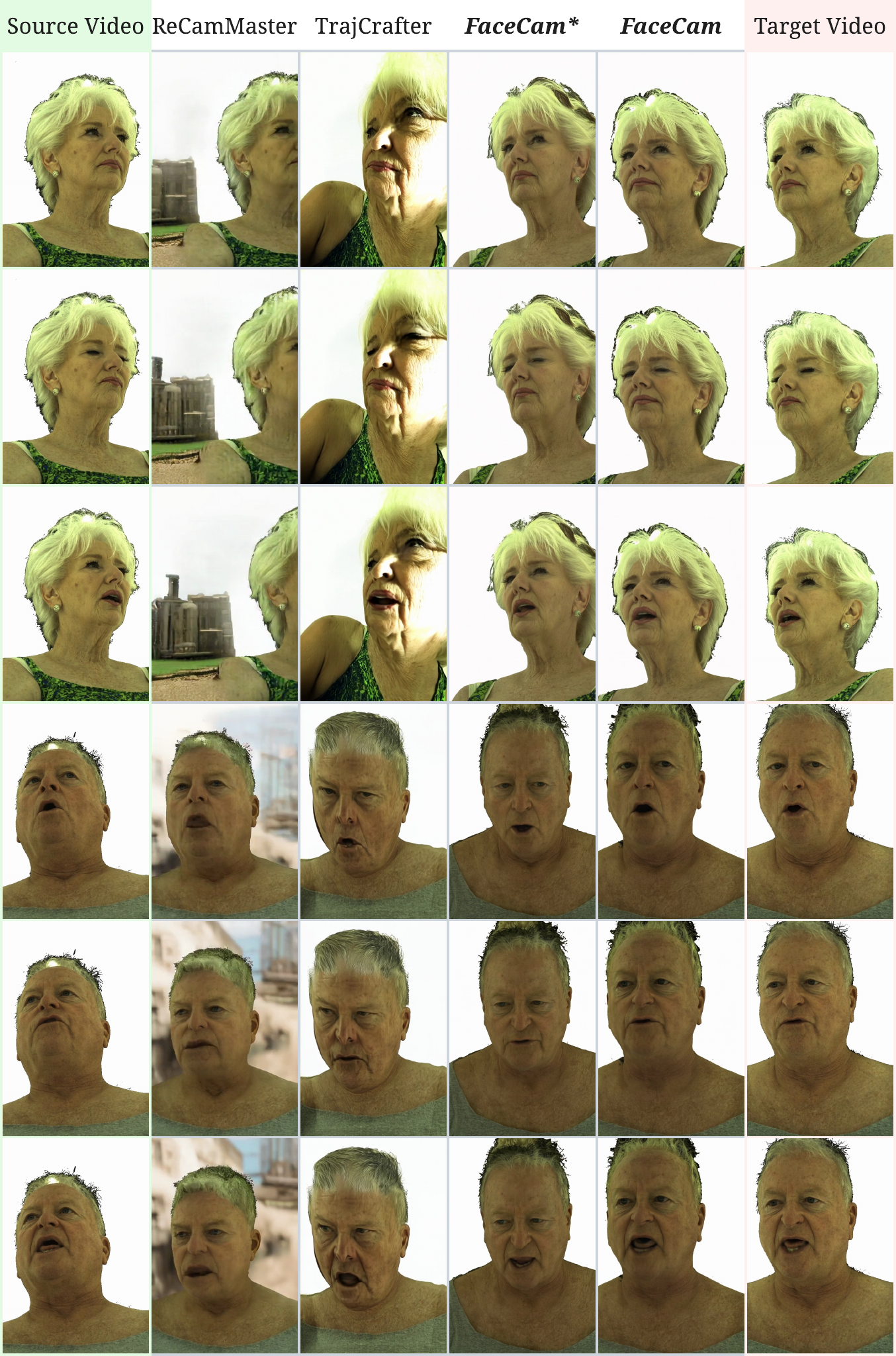}
    \vspace{-2mm}
    \caption{\textbf{Qualitative results on Ava-256.} {\ourmethod} produces more realistic, ground-truth-aligned novel views than baselines. ReCamMaster~\cite{recammaster} often fails under large pose changes, pushing the head out of frame, while TrajectoryCrafter~\cite{trajectorycrafter} frequently shows facial distortions from dynamic point-cloud errors.}
    \label{fig:ava256}
    \vspace{-4mm}
\end{figure}

\begin{table*}[t]
  \centering
  \scriptsize
  \caption{\textbf{Quantitative results on In-the-wild videos.} {\ourmethod} demonstrates superior identity preservation and camera trajectory correctness. It also generates videos with better visual quality and consistency as evidenced by VBench~\cite{vbench} scores.}
  \vspace{-2mm}
  \begin{tabular}{l@{\hskip 6.4mm}c@{\hskip 6.4mm}c@{\hskip 6.4mm}c@{\hskip 6.4mm}c@{\hskip 6.4mm}c@{\hskip 6.4mm}c@{\hskip 6.4mm}c@{\hskip 6.4mm}c}
    \toprule
    Method &
    \shortstack{Camera\\Correctness} &
    \shortstack{ArcFace\\Similarity} &
    \shortstack{Imaging\\Quality} &
    \shortstack{Aesthetic\\Quality} &
    \shortstack{Subject\\Consistency} &
    \shortstack{Background\\Consistency} &
    \shortstack{Motion\\Smoothness} &
    \shortstack{Dynamic\\Degree}\\
    \midrule
    ReCamMaster~\cite{recammaster} & 83.00 & 78.92 & 69.05 & 55.85 & 93.26 & 93.02 & \textbf{99.30} & 90.00 \\
    TrajectoryCrafter~\cite{trajectorycrafter} & 99.00 & 49.79 & 71.37 & 55.76 & 92.23 & 92.25 & 98.97 & \textbf{97.00} \\
    \ourmethod~(w/o In-the-wild Videos) & \textbf{100.00} & 77.73 & 70.71 &  55.73 & 94.52 & \textbf{95.16} & 99.23 & 89.00  \\
    \ourmethod & 97.00 & \textbf{83.94} & \textbf{73.49} & \textbf{59.91} & \textbf{94.77} & 94.98 & 99.05 & 96.00  \\
    \bottomrule
  \end{tabular}
  \label{tab:adobe_stock}
  \vspace{-4mm}
\end{table*}

\begin{figure*}[t]
    \centering
    \includegraphics[width=.98\linewidth]{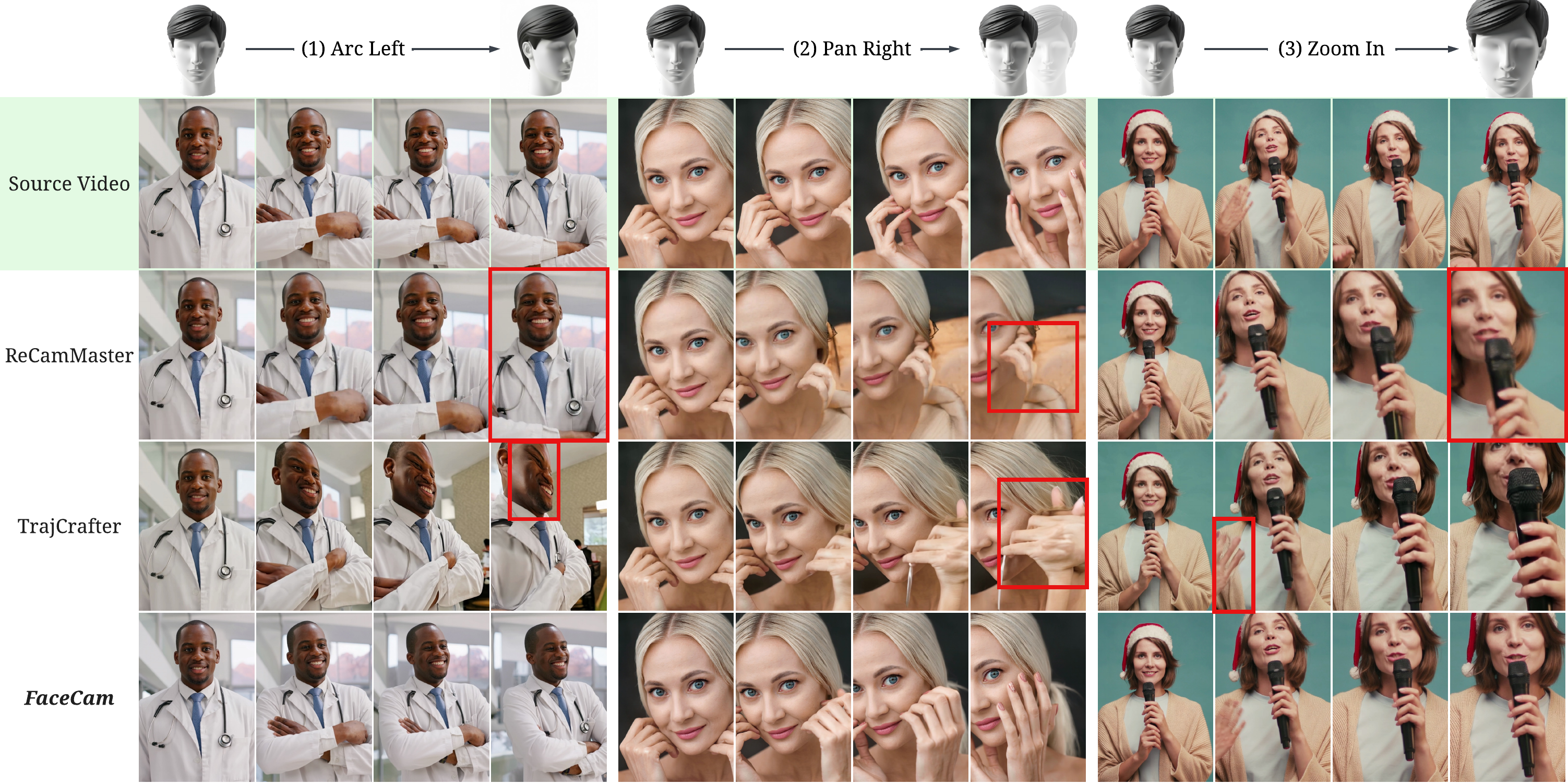}
    \vspace{-1mm}
    \caption{\textbf{Qualitative results on in-the-wild videos.} We present three camera motions: (1) Arc Left, (2) Pan Right, and (3) Zoom In. ReCamMaster~\cite{recammaster} often loses camera control in angle changes (panel 1) and produces blurry outputs under zoom in (panel 3). TrajectoryCrafter~\cite{trajectorycrafter} yields flattened faces with weak facial texture (panel 2). {\ourmethod} delivers higher visual quality and trajectory correctness, and more faithfully captures human geometry, including hands, hair, and facial features.}
    \label{fig:adobe_stock}
    \vspace{-1mm}
\end{figure*}
\vspace{-1mm}
\subsection{Experiments on Ava-256}
We report results on the Ava-256 dataset~\cite{ava256} in Tab.~\ref{tab:ava_256} and Fig.~\ref{fig:ava256}.  {\ourmethod} achieves stronger camera pose control and better identity and motion preservation than the baselines, demonstrating the benefit of our scale-aware camera representation and curated portrait training data. ReCamMaster~\cite{recammaster} often fails under large pose changes due to scale ambiguity, producing hallucinated backgrounds, and can only generate videos whose first frame matches the source. TrajectoryCrafter~\cite{trajectorycrafter} exhibits facial distortions due to errors in estimating and warping the dynamic point cloud, leading to lower identity preservation score.

\vspace{-1mm}
\subsection{Experiments on In-the-wild Portrait Videos}
Since ground-truth videos with known camera paths are unavailable, we evaluate camera-following accuracy via head-pose change. We use MediaPipe~\cite{mediapipe} to detect facial landmarks in the last frame of the generated and input videos, then estimate their head-pose difference. This pose change serves as a proxy for camera motion: for example, under a Pan Left target, the generated landmarks should shift right relative to the source; under an Arc Left target, the final head should face right relative to the source. We assign a binary correctness label to each video based on whether the measured pose shift matches the intended trajectory.

We present results in Tab.~\ref{tab:adobe_stock} and Fig.~\ref{fig:adobe_stock}. {\ourmethod} achieves high camera-motion correctness without explicit 3D geometry and attains stronger identity preservation than the baselines, highlighting the effectiveness of our scale-aware camera-conditioning design. ReCamMaster~\cite{recammaster} shows weaker camera control and often produces blur under zoom-in motions. TrajectoryCrafter~\cite{trajectorycrafter} relies on point-cloud estimation and lacks precise portrait-geometry reasoning, leading to lower ArcFace scores. Both baselines also struggle in outpainting (examples 2), whereas {\ourmethod} exhibits more robust portrait scene understanding.

We provide an ablation study on the training data in Tab.~\ref{tab:adobe_stock}. Training solely on NeRSemble~\cite{nersemble} without in-the-wild videos yields almost perfect camera movement correctness, but leads to lower identity preservation and image quality. Our full model achieves better identity preservation and visual quality, while maintaining high camera control adherence. More qualitative comparison and ablation studies are provided in the supplementary material.

\begin{figure*}[t]
    \centering
    \includegraphics[width=0.98\linewidth]{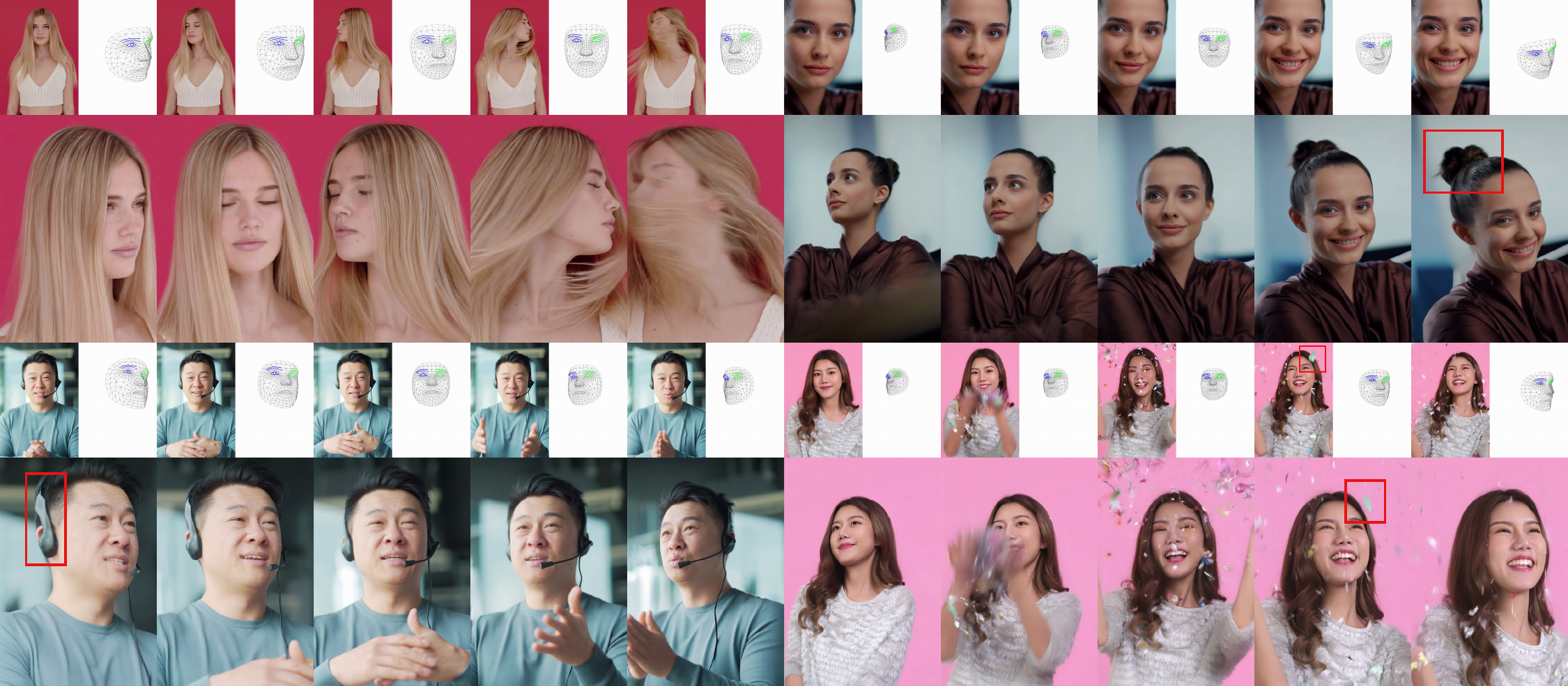}
    \vspace{-1mm}
    \caption{\textbf{In-the-wild results under diverse camera trajectories.} For each example, the first row shows the source video and the target camera, and the second row shows generated video.
    {\ourmethod} closely follows the specified trajectories and robustly handles everyday portrait scenarios: it synthesizes realistic outpainted regions when needed (\eg, bun hairstyles), preserves identity, expressions, and dynamic hair motion, and faithfully renders common co-occurring objects (\eg, headset). The example also highlights strong 3D understanding (\eg, flowing confetti).
    The first example further illustrates that the facial-landmark conditioning is not merely a representation of face location, but instead encodes camera pose and scale disentangled from head motion. \textbf{\textit{Zoom in for higher resolution and details.}}}
    \label{fig:random}
    \vspace{-1mm}
\end{figure*}

We further showcase {\ourmethod} under diverse, randomly sampled camera trajectories in Fig.~\ref{fig:random}, with varying azimuths, elevations, and FOVs. The results show robust performance across a wide range of motions and scenes, with strong preservation of facial features and expressions, consistent handling of human structure (\eg, hands and hair), and accurate synthesis of common co-occurring objects, indicating practical applicability in real-world settings.

\vspace{-1mm}
\section{Conclusion}
We introduce {\ourmethod}, a portrait video camera-control system that replaces scene-agnostic extrinsic camera representations with a face-tailored, scale-aware landmark representation. This conditioning resolves monocular scale ambiguity while providing intuitive, precise control over viewpoint. We further propose a data-generation pipeline that bootstraps from static multi-view studio captures and unlabeled in-the-wild videos via synthetic camera motion and multi-shot stitching, enabling continuous camera trajectories at inference without explicit 3D supervision. Experiments on Ava-256~\cite{ava256} and diverse in-the-wild videos demonstrate state-of-the-art camera controllability, stronger identity and motion preservation, and improved visual quality, validating both our representation and data strategy.

{
    \small
    \bibliographystyle{ieeenat_fullname}
    \bibliography{main}
}

\clearpage
\setcounter{section}{0}
\maketitlesupplementary

\section{Overview}
In this supplementary material, we first present a video (contains audio) overview of {\ourmethod} and its visual results. In Sec.~\ref{sec:supp_ablation}, we provide ablation studies on training data generation and proxy head selection. Additional qualitative results are shown in Sec.~\ref{sec:supp_results}. Implementation details are given in Sec.~\ref{sec:supp_implementation}. We include relevant preliminaries in Sec.~\ref{sec:supp_preliminary}, and discuss limitations and future work in Sec.~\ref{sec:supp_limitation}.

\section{Ablation Study}
\label{sec:supp_ablation}
\subsection{Training Data Generation}
\begin{table*}[t]
  \centering
  \scriptsize
  \caption{\textbf{Ablation study.} We conduct ablation studies to quantify the impact of different training data components on the final performance of our model. We also vary the choice of proxy head and show that this selection has negligible effect on the generated results.}
  \begin{tabular}{l@{\hskip 5.6mm}c@{\hskip 5.6mm}c@{\hskip 5.6mm}c@{\hskip 5.6mm}c@{\hskip 5.6mm}c@{\hskip 5.6mm}c@{\hskip 5.6mm}c@{\hskip 5.6mm}c}
    \toprule
    Method &
    \shortstack{Camera\\Correctness} &
    \shortstack{ArcFace\\Similarity} &
    \shortstack{Imaging\\Quality} &
    \shortstack{Aesthetic\\Quality} &
    \shortstack{Subject\\Consistency} &
    \shortstack{Background\\Consistency} &
    \shortstack{Motion\\Smoothness} &
    \shortstack{Dynamic\\Degree}\\
    \midrule
    \ourmethod~(w/o Synthetic Camera Motion) & 96.00 & 81.19 & 72.03 & 58.02 & 94.27 & 94.91 & 99.29 & 83.00 \\
    \ourmethod~(w/o Multi-shot Stitching) & 86.00 & 76.38 & 70.73 & 55.10 & 94.56 & 95.12 & \textbf{99.30} & 80.00 \\
    \ourmethod~(w/o In-the-wild Videos) & \textbf{100.00} & 77.73 & 70.71 &  55.73 & 94.52 & \textbf{95.16} & 99.23 & 89.00  \\
    \ourmethod & 97.00 & \textbf{83.94} & \textbf{73.49} & \textbf{59.91} & \textbf{94.77} & 94.98 & 99.05 & \textbf{96.00}  \\
    \midrule[0.1pt]  
    \ourmethod~(Proxy 3D Head 1) & 97.00 & 84.45 & 73.48 & 59.85 & 94.80 & 94.89 & 99.02 & 95.00  \\
    \ourmethod~(Proxy 3D Head 2) & 97.00 & 84.74 & 73.47 & 59.89 & 94.74 & 94.89 & 99.03 & 94.00   \\
    \bottomrule
  \end{tabular}
  \label{tab:supp_ablation}
\end{table*}

\begin{figure*}[b]
    \centering
    \includegraphics[width=0.98\linewidth]{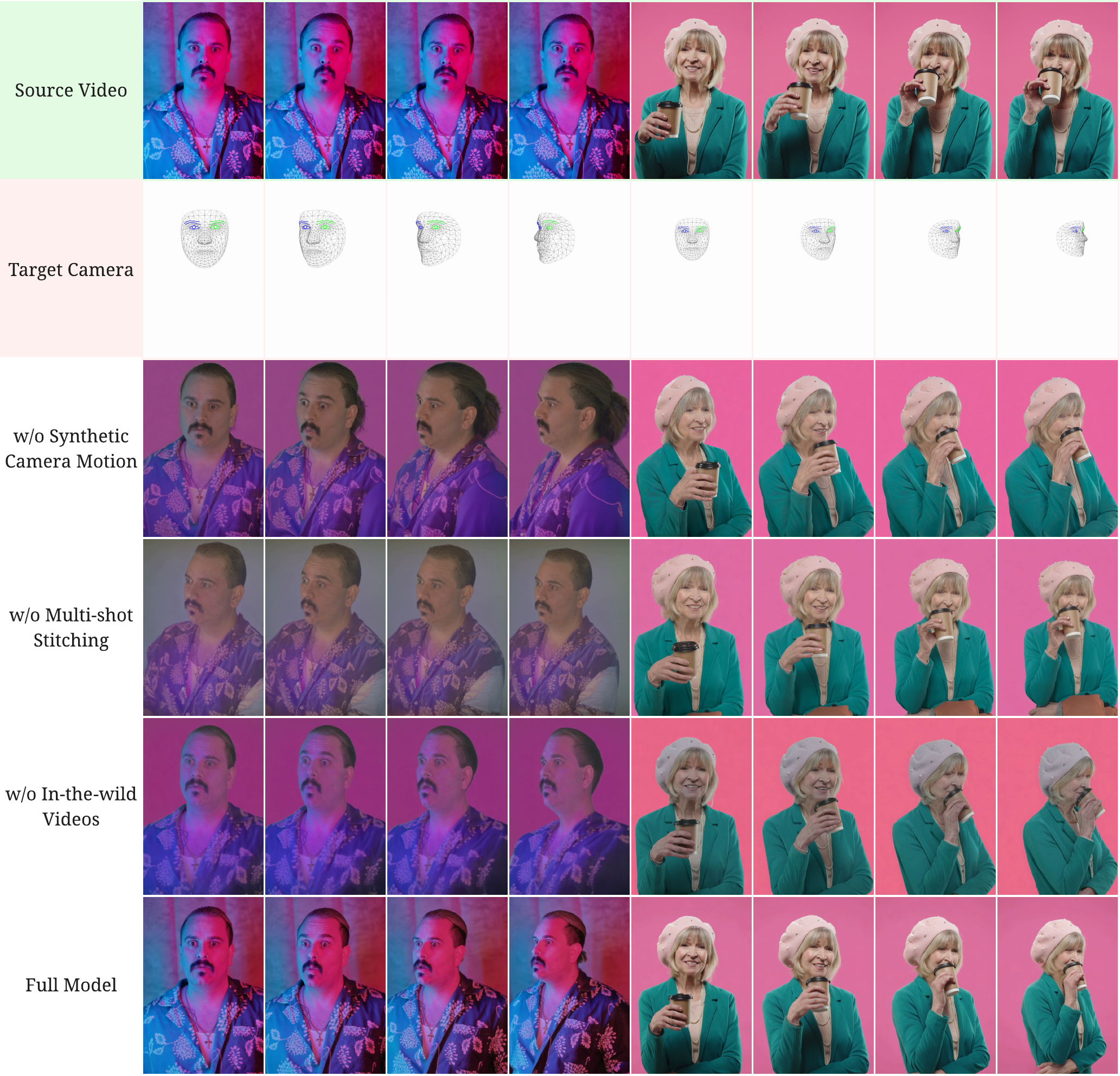}
    \caption{\textbf{Ablation study on training data generation.} Without Synthetic Camera Motion, the model often produces inaccurate camera trajectories with discontinuous or abrupt changes. Without Multi-shot Stitching, the model cannot learn to change camera angles along a trajectory. With both strategies applied but without in-the-wild videos (\textit{w/o In-the-wild Videos}), the model generates correct camera motion and angle changes, but the lighting remains tied to the training distribution and fails to generalize to real-world illumination, leading to inconsistencies with the source video. Our full model provides accurate camera control and high image quality with lighting and appearance consistent with the source video.}
    \label{fig:supp_ablation}
\end{figure*}

We provide an ablation study on training data generation in Tab.~\ref{tab:supp_ablation} and Fig.~\ref{fig:supp_ablation} to analyze the impact of each strategy on the final results. We compare three ablated variants and our full model:

\begin{itemize}
\item \textit{{\ourmethod} (w/o Synthetic Camera Motion)}: applies only Multi-shot Stitching to NeRSemble~\cite{nersemble} videos.
\item \textit{{\ourmethod} (w/o Multi-shot Stitching)}: applies only Synthetic Camera Motion to NeRSemble videos.
\item \textit{{\ourmethod} (w/o In-the-wild Videos)}: applies both Synthetic Camera Motion and Multi-shot Stitching to NeRSemble videos, without using any in-the-wild videos.
\item \textit{{\ourmethod}}: our full model, which adds in-the-wild videos with Synthetic Camera Motion (since multi-view videos are not available) on top of the third baseline.
\end{itemize}

Through these experiments, we observe that Synthetic Camera Motion enables the model to learn zoom and pan motions and to produce smooth trajectories without sudden camera pose changes. Multi-shot Stitching further teaches the model to follow camera angle changes along the target trajectory, and together these two strategies yield accurate camera control. Incorporating in-the-wild video data improves generalization to diverse real-world lighting conditions and objects, leading to better appearance consistency with the input video.

\subsection{Proxy 3D Head Selection}
\begin{figure}[t]
    \centering
    \includegraphics[width=.98\linewidth]{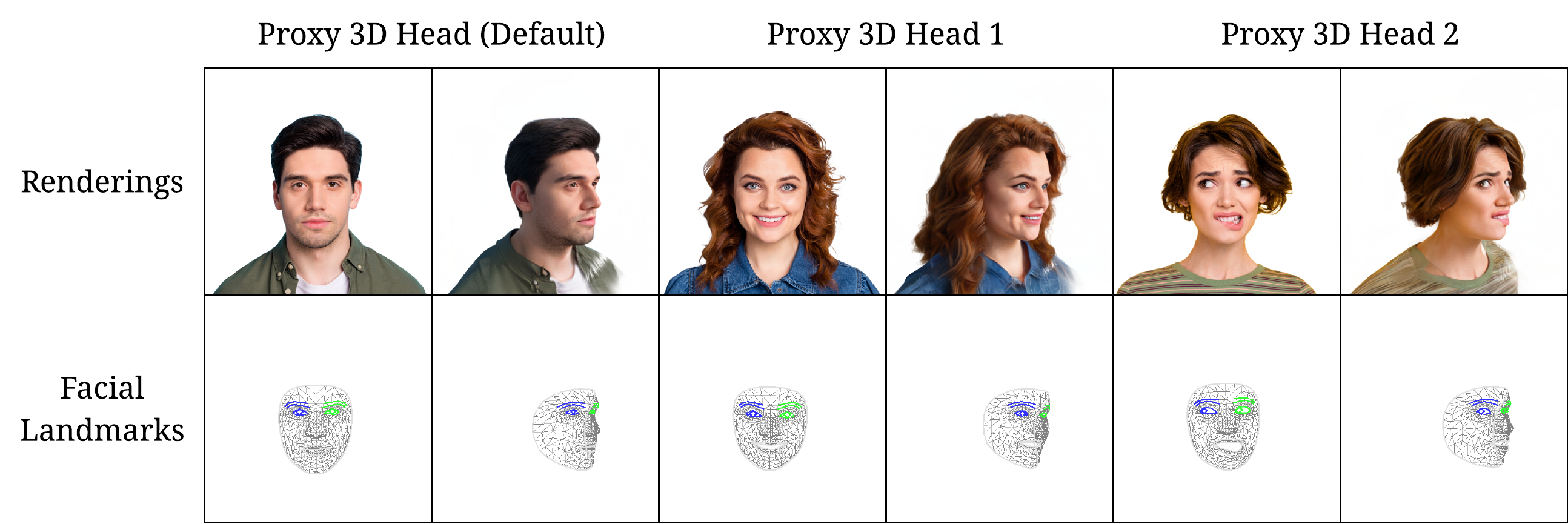}
    \caption{\textbf{Different choices of proxy 3D head}. We select two additional proxy 3D heads with different identities and corresponding facial landmark detections, and conduct an ablation study showing that the proxy’s identity and expression do not affect the final generation results.}
    \label{fig:supp_ablation_proxy}
\end{figure}
For all experiments reported in the main paper, we use a single generic 3D Gaussian head as a proxy during inference to render videos and extract facial landmarks. This proxy head can be generated by any 3D head generation methods~\cite{facelift, avat3r}. To verify that the specific choice of proxy head does not influence performance, we select two additional proxy heads (Fig.~\ref{fig:supp_ablation_proxy}) and evaluate {\ourmethod} on in-the-wild videos. The results in Tab.~\ref{tab:supp_ablation} show only minor differences across all three proxies, indicating that our approach is largely insensitive to the particular proxy head. This supports our design choice that the landmarks serve purely as a camera-conditioning signal, rather than conveying identity or expression information. The identity and expression in the generated videos come solely from the source video.

\section{Experimental Results}
\label{sec:supp_results}
\begin{figure*}[tb]
    \centering
    \includegraphics[width=0.8\linewidth]{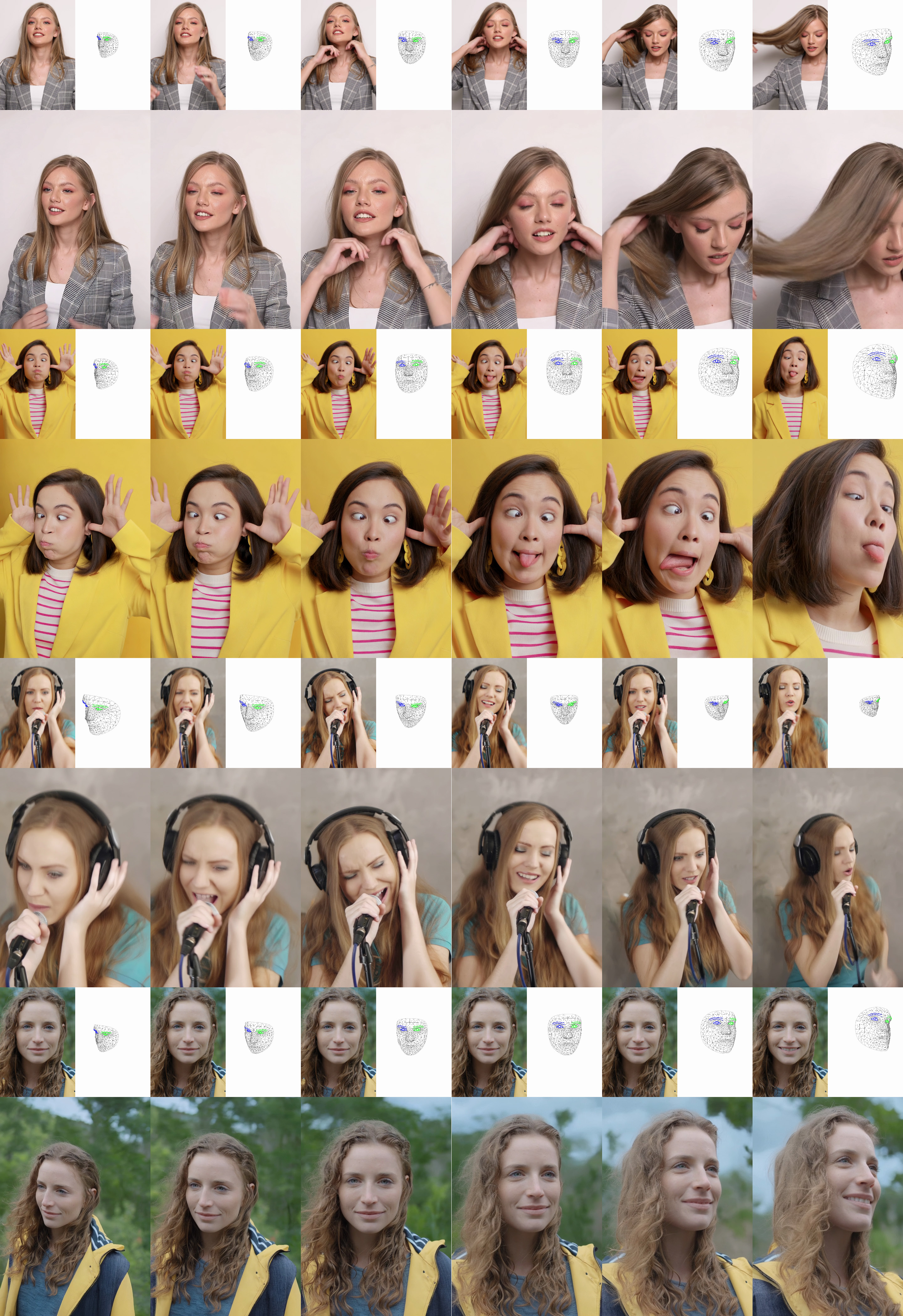}
    \caption{{\ourmethod} performs robustly across diverse, challenging scenarios: it retains head and hair motion from the input video (example 1), captures a wide range of facial expressions (example 2), maintains motion-induced blur from fast body movements (example 3), and plausibly outpaints clothing and background when the generated video contains a smaller face region than the input (example 4).}
    \label{fig:supp_random_1}
\end{figure*}

\begin{figure*}[tb]
    \centering
    \includegraphics[width=0.8\linewidth]{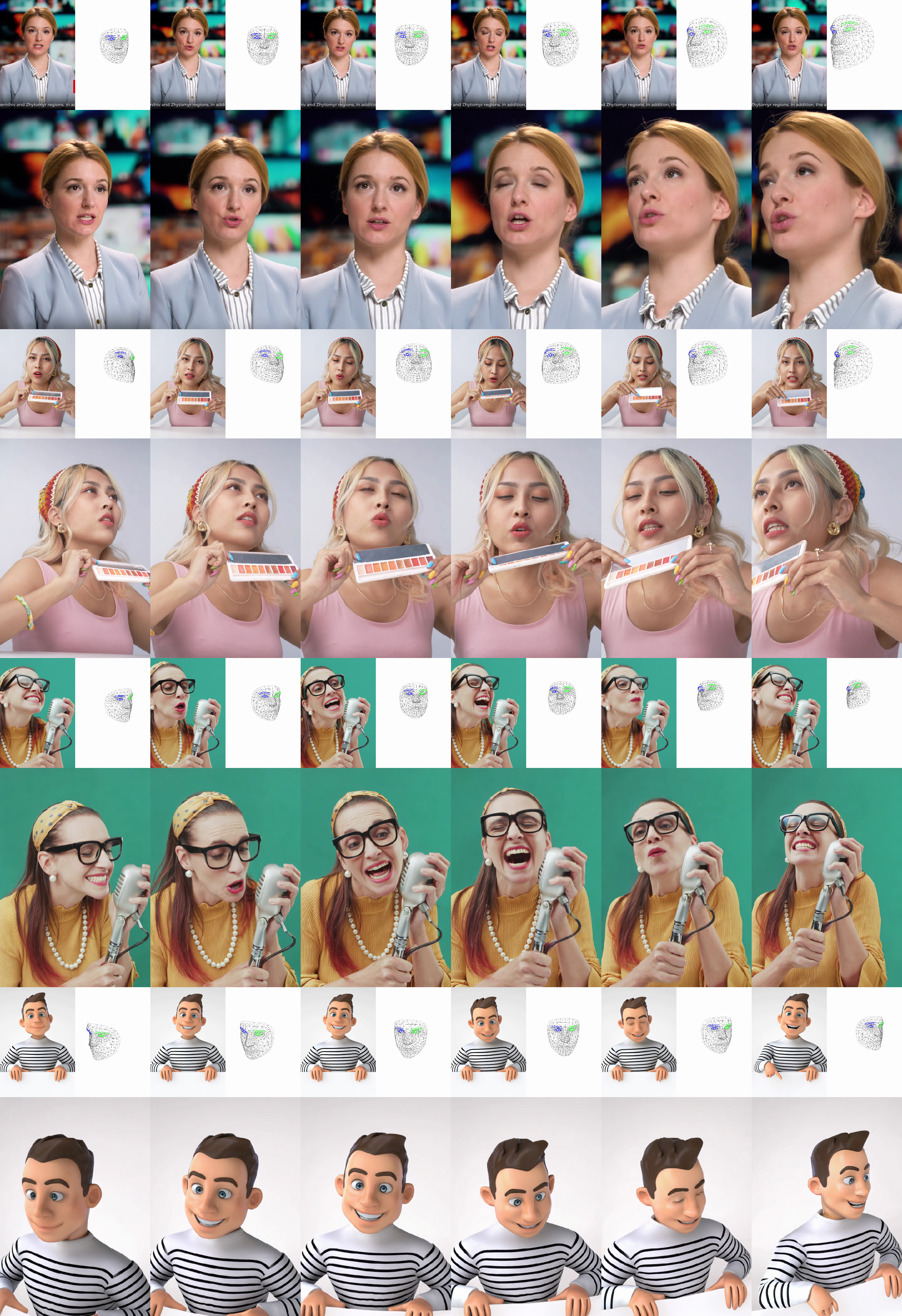}
    \caption{{\ourmethod} in real-world scenarios. It recaptures a newscaster with detailed facial texture while keeping the studio background consistent (example 1). It further re-synthesizes an e-commerce streamer (example 2) and a singer (example 3) under novel camera angles, accurately maintaining co-occurring objects such as an eyeshadow palette, earrings, headband, necklace, microphone, and glasses, \etc. The model even generalizes to cartoon characters (example 4), despite never having seen such content during training.}
    \label{fig:supp_random_2}
\end{figure*}

We conduct extensive experiments on in-the-wild videos with diverse camera trajectories to assess {\ourmethod} under both challenging synthetic settings and realistic use cases. The results are shown in Fig.~\ref{fig:supp_random_1} and Fig.~\ref{fig:supp_random_2}. Across a wide range of inputs, the model tracks intricate head and hair dynamics, responds smoothly to varied facial expressions, and respects motion-dependent artifacts such as blur from rapid body movement. When the target trajectory places the virtual camera farther from the subject, {\ourmethod} plausibly completes missing regions by synthesizing coherent clothing and background content. On real footage, it re-creates studio and streaming scenes with stable identity and layout, while reliably retaining fine-grained accessories and props (\eg., cosmetics, jewelry, headbands, microphones, and glasses). Notably, the same pipeline extends to stylized inputs such as cartoon characters, indicating strong generalization beyond the distribution of the training data.

\section{Implementation Details}
\label{sec:supp_implementation}
\subsection{Training Data Generation}
\begin{algorithm}[t]
\caption{Scale and Color Augmentation}
\label{alg:scale_color_aug}
\begin{algorithmic}[1]
\Require Source clip $V^s = \{I^s_i\}_{i=1}^{T_s}$, target clip $V^t = \{I^t_i\}_{i=1}^{T_t}$
\Ensure Augmented clips $\tilde{V}^s, \tilde{V}^t$
\State Sample scale factors $s^s, s^t \sim \mathcal{U}(0.75, 1.25)$
\State Sample a background color $c \sim \text{UniformColor()}$ \Comment{shared between source and target}
\vspace{0.5mm}
\For{$i = 1$ to $T_s$} \Comment{augment source clip}
    \State $J^s_i \gets \text{Resize}(I^s_i, s^s)$
    \State $M^s_i \gets \text{FaceSeg}(J^s_i)$ \Comment{$M^s_i \in \{0,1\}^{H \times W}$}
    \State $B^s_i \gets c \cdot (\mathbf{1} - M^s_i)$
    \State $\tilde{I}^s_i \gets M^s_i \odot J^s_i + B^s_i$
\EndFor
\State $\tilde{V}^s \gets \{\tilde{I}^s_i\}_{i=1}^{T_s}$
\vspace{0.5mm}
\For{$i = 1$ to $T_t$} \Comment{augment target clip with same background color}
    \State $J^t_i \gets \text{Resize}(I^t_i, s^t)$
    \State $M^t_i \gets \text{FaceSeg}(J^t_i)$
    \State $B^t_i \gets c \cdot (\mathbf{1} - M^t_i)$
    \State $\tilde{I}^t_i \gets M^t_i \odot J^t_i + B^t_i$
\EndFor
\State $\tilde{V}^t \gets \{\tilde{I}^t_i\}_{i=1}^{T_t}$
\end{algorithmic}
\end{algorithm}

\begin{algorithm}[t]
\caption{Synthetic Camera Motion (Zoom and Pan)}
\label{alg:synthetic_motion}
\begin{algorithmic}[1]
\Require Input clip $V = \{I_i\}_{i=1}^{T}$, motion type $m \in \{\text{zoom}, \text{pan}\}$, image resolution $(H, W)$
\Ensure Motion-augmented clip $\tilde{V}$
\vspace{0.5mm}
\If{$m = \text{zoom}$}
    \State Sample $s_{\text{start}}, s_{\text{end}} \sim \mathcal{U}(1.0, 1.25)$
    \For{$i = 1$ to $T$}
        \State $\alpha \gets \frac{i-1}{\max(T-1, 1)}$
        \State $s_i \gets (1 - \alpha) \cdot s_{\text{start}} + \alpha \cdot s_{\text{end}}$
        \State $J_i \gets \text{Resize}(I_i, s_i)$
        \State $\tilde{I}_i \gets \text{CenterCropOrPad}(J_i, H, W)$
    \EndFor
\ElsIf{$m = \text{pan}$}
    \State Choose maximum offset $\delta_x, \delta_y$ relative to $(H, W)$
    \State Sample offsets $\mathbf{o}_{\text{start}}, \mathbf{o}_{\text{end}} \sim [-\delta_x, \delta_x] \times [-\delta_y, \delta_y]$
    \For{$i = 1$ to $T$}
        \State $\alpha \gets \frac{i-1}{\max(T-1, 1)}$
        \State $\mathbf{o}_i \gets (1 - \alpha)\,\mathbf{o}_{\text{start}} + \alpha\,\mathbf{o}_{\text{end}}$
        \State $\tilde{I}_i \gets \text{CropOrPadWithOffset}(I_i, \mathbf{o}_i, H, W)$
    \EndFor
\EndIf
\State $\tilde{V} \gets \{\tilde{I}_i\}_{i=1}^{T}$
\end{algorithmic}
\end{algorithm}

\begin{algorithm}[t]
\caption{Multi-shot Stitching}
\label{alg:multishot_stitching}
\begin{algorithmic}[1]
\Require Set of clips for a target video $\mathcal{C} = \{V^{(k)}\}_{k=1}^{N}$, $V^{(k)} = \{I^{(k)}_i\}_{i=1}^{T_k}$, maximum shots $K_{\max} = 4$
\Ensure Stitched target clip $\tilde{V}$
\vspace{0.5mm}
\State Sample number of shots $K \sim \text{Uniform}\{1, 2, \dots, K_{\max}\}$
\State Sample $K$ distinct indices $\{i_1, \dots, i_K\}$ from $\{1, \dots, N\}$ \Comment{different camera poses}
\State Initialize stitched sequence $\tilde{V} \gets \emptyset$
\vspace{0.5mm}
\For{$j = 1$ to $K$}
    \State $V^{(j)} \gets V^{(i_j)}$, length $T^{(j)}$
    \State Sample start index $a_j \sim \{1, \dots, T^{(j)} - 1\}$
    \State Sample end index $b_j \sim \{a_j + 1, \dots, T^{(j)}\}$
    \State Define segment $S_j \gets \{ I^{(j)}_i \mid i = a_j, \dots, b_j \}$
    \State $\tilde{V} \gets \text{Concat}(\tilde{V}, S_j)$
\EndFor
\State \Return $\tilde{V}$
\end{algorithmic}
\end{algorithm}

We provide pseudo-code for three training data generation procedures. \textit{Scale and Color Augmentation} (Algorithm~\ref{alg:scale_color_aug}) is applied to both source and target videos to increase data diversity, including variations in head size and background appearance. \textit{Synthetic Camera Motion} (Algorithm~\ref{alg:synthetic_motion}) is applied to the target video to create continuous camera trajectories with zoom and pan effects, which is essential for achieving smooth, temporally coherent generated videos. Finally, \textit{Multi-shot Stitching} (Algorithm~\ref{alg:multishot_stitching}) is applied to the target video to introduce discrete camera pose changes, enabling the model to handle viewpoint transitions in the generated outputs.


\begin{table}[t]
    \centering
    \caption{Source and target camera pairs used in experiment on Ava-256~\cite{ava256}.}
    \begin{tabular}{lcc}
        \toprule
        \textbf{ID} & \textbf{Source Camera} & \textbf{Target Camera} \\
        \midrule
        1  & cam\_400944 & cam\_401031 \\
        2  & cam\_400944 & cam\_401410 \\
        3  & cam\_400981 & cam\_401045 \\
        4  & cam\_400981 & cam\_401292 \\
        5  & cam\_401163 & cam\_401031 \\
        6  & cam\_401163 & cam\_401458 \\
        7  & cam\_401168 & cam\_401045 \\
        8  & cam\_401168 & cam\_401292 \\
        9  & cam\_401316 & cam\_401410 \\
        10 & cam\_401316 & cam\_401458 \\
        \bottomrule
    \end{tabular}
    \label{tab:supp_cam_pairs}
\end{table}

\subsection{Experimental Details}
All experiments are conducted at a resolution of $704 \times 480$, with generated videos of length 81 frames. We use the same text prompt for all experiments: ``A portrait of a person.'' 
TrajectoryCrafter~\cite{trajectorycrafter} can generate at most 49 frames in the general setting, and only 29 frames when the first frame of the generated video does not coincide with the first frame of the source video. To ensure a fair comparison, we therefore evaluate on the first 29 frames in the static-camera setting and on the first 49 frames in the dynamic-camera setting for all baselines.
ReCamMaster~\cite{recammaster} produces camera-controlled results only when the target camera pose for the first frame has an identity rotation; otherwise, the generated video degenerates to the source video. In the static-camera experiments, we thus enforce an identity rotation as the first-frame camera condition for this baseline to obtain valid results.
All baselines are run with their official configurations and released pre-trained weights.

For the static-camera setting on the Ava-256 dataset, we select 10 identities, each with 10 source–target camera pairs, yielding a total of 100 videos.
The selected identities are \texttt{KDA058}, \texttt{XJT672}, \texttt{LAS440}, \texttt{IFG774}, \texttt{EID363}, \texttt{NRE683}, \texttt{PAK800}, \texttt{MCR809}, \texttt{SKB942}, \texttt{KJJ701}. The source and target cameras are summarized in Tab.~\ref{tab:supp_cam_pairs}.
\section{Preliminary}
\label{sec:supp_preliminary}
\subsection{Conditional Video Generation}
We build our system on the open-source video foundation model Wan~\cite{wan2025} for conditional video generation. Wan is a latent video diffusion model comprising a 3D Variational Autoencoder (VAE)~\cite{vae}, a text prompt encoder~\cite{t5}, and two transformer-based diffusion models (DiT)~\cite{dit} specialized for the high and low noise stages. The model adopts Rectified Flow framework~\cite{rectified_flow_transformers} for the noise schedule and denoising process. Detailed architecture and training settings are provided in the supplementary material.

During training, a pre-trained 3D VAE encodes a video $V\in\mathbb{R}^{f\times h\times w\times c}$ into latent space: $z_0=\mathcal{E}(V)$.
Then in the forward diffusion process, the DiT injects Gaussian noise $\epsilon$ into $z_0$ to create a noisy latent. The forward process is defined as straight paths between data distribution and a standard normal distribution.
\begin{equation}
    z_t = (1-t)z_0 + t\epsilon,\qquad \varepsilon\sim\mathcal{N}(0,I),
\end{equation}
where $t$ denotes the iterative timestep. $z_t$ is then patchified, concatenated with text tokens encoded by the text prompt encoder and other additional conditioning signals, and fed into DiT blocks.
To solve the reverse denoising process, Conditional Flow Matching (CFM)~\cite{flow_matching} learns a time-dependent velocity field $v_{\theta}(z,t,\mathbf{c})$ that defines an ordinary differential equation (ODE):
\begin{equation}
    \frac{dz_t}{dt}=v_{\theta}(z_t,t,\mathbf{c}),\qquad t\in[0,1],
\end{equation}
transporting samples from the base (standard Gaussian) to the data distribution under conditioning $\mathbf{c}$. With the rectified interpolant, the target velocity along the path is constant:
\begin{equation}
    u^\star(z_t,t|z_0,\varepsilon)=\frac{d z_t}{dt}=\varepsilon - z_0 .
\end{equation}
CFM trains $v_{\theta}$  by regressing to this target with MSE loss:
\begin{equation}
    \mathcal{L}_{\text{CFM}} =\mathbb{E}_{t, z_0, \epsilon} \bigl|\bigl|v_{\theta}(z_t,t,\mathbf{c})-(\varepsilon-z_0)\bigr|\bigl|_2^2 .
\end{equation}
At inference, we integrate the learned ODE deterministically from noise to data by marching from $t=1$ to $t=0$:
\begin{equation}
    z_{t-\Delta t}=z_t-\Delta t~v_{\theta}(z_t,t,\mathbf{c}),
\end{equation}
yielding the final latent $z_0$ consistent with the conditioning $\mathbf{c}$. $z_0$ is then decoded by the pre-trained VAE decoder and outputs the generated video: $V = \mathcal{D}(z_0)$.

\subsection{Wan2.2 and MoE Video Diffusion}
Wan2.2~\cite{wan2025} is a family of large-scale latent video diffusion models. It supports multi-modal conditioning (text-to-video, image-to-video, text–image-to-video, and specialized speech/animation variants) and generates high-fidelity videos up to 720p at 24\,fps using a high-compression video VAE~\cite{vae} and a DiT-style~\cite{dit} diffusion backbone.

To scale model capacity without increasing inference cost, Wan2.2 replaces a single denoising network with a Mixture-of-Experts (MoE)~\cite{wan2025wan22} architecture. A set of expert denoisers is specialized for different noise regimes (\eg., high-noise early steps \vs low-noise late steps), and a routing scheme based on the diffusion timestep selects which expert to apply at each step. This design enlarges the total parameter count and improves motion, semantic, and aesthetic fidelity, while keeping per-step FLOPs comparable to a dense model.

More generally, an MoE layer consists of a collection of experts $\{E_k\}_{k=1}^K$ and a gating function $g(x)$ that selects a sparse subset of experts for each input $x$, often via top-$k$ routing. Only the selected experts are evaluated and their outputs are combined, for example
\begin{equation}
    y \;=\; \sum_{k \in \mathcal{S}(x)} g_k(x)\,E_k(x),
\end{equation}
where $\mathcal{S}(x)$ is a small set of active experts and $g_k(x)$ are normalized routing weights. By activating only a few experts per input, MoE architectures enable models with billions of parameters to operate at roughly the same compute cost as much smaller dense networks, a property that Wan2.2 leverages to scale video generation quality and controllability.

\subsection{MediaPipe Facial Landmark Detection}
We use Google’s MediaPipe Face Mesh~\cite{mediapipe} as an off-the-shelf module to obtain dense 2D/3D facial keypoints from monocular RGB inputs. MediaPipe Face Mesh predicts a set of $K=468$ landmarks in real time, even on mobile devices, by applying a lightweight neural network to a cropped face region and regressing per-vertex coordinates that approximate the full facial surface. The model operates on a single RGB camera without requiring depth sensors and is optimized for GPU acceleration, making it suitable for large-scale video processing and interactive applications.

Concretely, given an input frame $I_i$, the detector returns a landmark set
\begin{equation}
    \mathbf{U}_i = \{\mathbf{u}_{i,k}\}_{k=1}^{K}, \quad K = 468,
\end{equation}
where each
\begin{equation}
    \mathbf{u}_{i,k} = (x_{i,k}, y_{i,k}, z_{i,k})
\end{equation}
encodes normalized image coordinates $(x_{i,k}, y_{i,k})$ and a relative depth value $z_{i,k}$. In practice, MediaPipe adopts a two-stage pipeline: a BlazeFace-style face detector first produces a tight region of interest, and a dedicated mesh regressor then predicts the dense landmark configuration within that region. These landmarks are widely used in AR and avatar applications to recover facial geometry and pose from video streams; in our work, we reuse them as a compact, robust representation for conditioning and camera control.

\section{Limitations and Future Work}
\label{sec:supp_limitation}
Despite the accurate camera control and high-quality results achieved, {\ourmethod} still has several limitations.
First, because facial landmarks can only be detected when facial features are visible in the input video, {\ourmethod} cannot handle views where the camera rotates to the back of the head. For the same reason, although {\ourmethod} can generalize to cartoon characters, it does not extend to general scenes in which a facial landmark detector is inapplicable. Building on the same idea of using image-space correspondences as a camera representation, but redefining how these correspondences are encoded, could help address this limitation.
Second, background generation is not the focus of this work, partly due to data limitations. Incorporating synthetic data with multi-view-consistent backgrounds could further improve the model’s ability to synthesize background content behind the subject.
Third, due to the limitations of the underlying video generation model, {\ourmethod} remains relatively slow at inference and is not yet suitable for real-time applications. Distilling the model or adopting a more efficient video generation backbone are promising directions.


\clearpage

\end{document}